\documentclass[preprint]{elsarticle}

\usepackage{amssymb}
\usepackage{mathabx}
\usepackage{mathdots}
\usepackage{algpseudocode}
\usepackage{algorithm}
\usepackage{bbm}
\usepackage{graphicx}
\usepackage{multirow}
\usepackage{wasysym}
\usepackage{latexsym}
\usepackage{index}
\usepackage{forest}
\usepackage{color}
\usepackage{url}
\usepackage{hyperref}

\definecolor{SeaGreen}{HTML}{2E8B57}
\definecolor{myGreen}{HTML}{66CC00}
\definecolor{Maroon}{HTML}{800000}
\definecolor{myGrey}{HTML}{A0A0A0}
\definecolor{LstItemCol}{HTML}{3333FF}
\usepackage[nottoc]{tocbibind}

\begin{document}

\begin{frontmatter}

\title{A Framework of Transfer Learning in Object Detection for Embedded Systems}

\author[rvt]{Ioannis Athanasiadis}
\ead{asioanni@ece.auth.gr}

\author[rvt]{Panagiotis Mousouliotis\corref{cor1}}
\ead{pmousoul@ece.auth.gr}

\author[rvt]{Loukas Petrou}
\ead{loukas@eng.auth.gr}

\address[rvt]{School of Electrical and Computer Engineering, Aristotle University of Thessaloniki, 54124 Thessaloniki, Greece}
\cortext[cor1]{Corresponding author. Tel: +30 6974 216472}

    \begin{abstract}
    Transfer learning is one of the subjects undergoing intense study in the area of machine learning. In object recognition and object detection there are known experiments for the transferability of parameters, but not for neural networks which are suitable for object detection in real time embedded applications, such as the SqueezeDet neural network. We use transfer learning to accelerate the training of SqueezeDet to a new group of classes. Also, experiments are conducted to study the transferability and co-adaptation phenomena introduced by the transfer learning process. To accelerate training, we propose a new implementation\footnote{The repository with the framework’s implementation and the experiments: \url{https://github.com/supernlogn/squeezeDetTL}} of the SqueezeDet training which provides a faster pipeline for data processing and achieves 1.8 times speedup compared to the initial implementation. Finally, we created a mechanism for automatic hyperparameter optimization using an empirical method.

    \end{abstract}

    
\end{frontmatter}

\section{Introduction}
\label{sect:Introduction}
Retraining a convolutional neural network (CNN) to a new object detection dataset is usually a hard and time consuming task requiring an expert on the field to handle the retraining procedure. Nevertheless, there are numerous embedded applications where the object detection task is constantly changing by requiring the detection of new classes of objects. A typical example is an autonomous car driving and tracking objects with its camera in real time. The car should be able to track new objects when its manufacturer requires it. Another example is that of a microscope which uses object detection to track blood cells; it should be able to track new cells when the medical staff requires it. There are many relevant examples which all come down to the same architecture: a database,  which holds the parameters of an object detection model and which is able to be updated fast and easily; this is where Transfer Learning comes into play.

Transfer learning (TL) is the method of using knowledge from a previous training to a new training aiming to make possible the training in a new dataset and accelerate the process \cite{40}. In the context of CNNs, transfer learning can be implemented by transplanting the learned feature layer parameters from one CNN (derived from the source task) to initialize another layer (for the target task). There are other types of TL in neural networks which are discussed below, but the most basic is the one where parameter transfer takes place.

Transferability is the most important metric of TL. It uses a metric of training (e.g. accuracy) and measures the difference between the metric of a training with TL and without TL. If the difference is positive then the transferability is positive, else it is negative. Apparently, in any application transferability should be positive. From the view of transferability, researchers in \cite{55} have proposed an experiment to study the transferability of the layers of AlexNet \cite{20}. AlexNet is used for image recognition which is a common field for TL experiments.

However, in the field of object detection there are no separate experiments on the parameter transferability and on the fragility to co-adaptation. In object detection, more processing is required by the layers following the feature extractor, which is typically a CNN, than in object recognition where the feature extractor is followed by a classifier. Object detection neural networks should also predict the position and the class of the object inside an image. This means that more information should be carried from the previous layers to the last ones. This last observation leads to the assumption that CNN layers which are used for object detection are more fragile to co-adaptation. To validate this assumption, an experiment is performed using KITTI \cite{67} and PASCAL VOC \cite{68} datasets. It is evident that the transferability depends on the choice of the source dataset \cite{102}.

Moreover, having in mind embedded applications such as the two mentioned above, the experiment should be performed on a CNN suitable for embedded devices. For this purpose, experiments of co-adaptation and transferability used SqueezeDet \cite{1}. SqueezeDet is one of the smallest neural networks for object detection and it is based on SqueezeNet which can be used in real time applications \cite{5,103,104}. The co-adaptation experiment from KITTI to PASCAL VOC revealed that SqueezeDet layers are very fragile to this phenomenon and also experiments from ImageNet to PASCAL VOC showed that TL can help the training of SqueezeDet in a small dataset achieve 40+\% accuracy which would not be possible without TL. This leads to the conclusion that the transferability metric of this retraining using the SqueezeDet network is highly positive.

The vanilla version of SqueezeDet's training uses the CPU heavily, making it a very time consuming process. To overcome this problem, the training was redesigned and reimplemented to use GPU acceleration, speeding up the process by $1.8\times$ compared to the vanilla training implementation. During the re-implementation we built a new framework which can optimize the hyperparameters of the neural network for the specific dataset automatically. This easies the update of a database containing network parameters.

In the following we first present the background of this field in section \ref{sect:Background}. We develop our methodology in section \ref{sect:Methods}, to solve the problems of retraining SqueezeDet and running transfer learning experiments with optimal hyperparameters. Finally, the experiments arising from the methodology are presented in section \ref{sect:Experiments} and we provide our conclusions in section \ref{sect:Conclusions}.

\section{Background}
\label{sect:Background}
\subsection{Parameter Transfer on CNNs}
\label{subsect:Parameter Transfer on CNNs}
Transfer Learning by parameter transfer is the easiest and most common way to transfer knowledge and, as such, it has been studied in more detail. CNNs can be described as stacks of layers that extract features; the first layers extract more general features. Specifically for images, the features are common and even an added SVM at the top is enough to produce acceptable results for a new dataset \cite{90}. Usually, the first layers are Gabor filters \cite{55} for edge and line detection.  Moving from the input to the final classifier more specific features of the dataset are discovered. The features extracted from the first layers of a CNN are much more likely to be found in other datasets than those extracted by the latter layers \cite{90}.

\subsection{Transfer Learning in Object Detection}
In a CNN used for recognition, all layers have as common purpose to extract features about the object class. On the other hand, when a CNN is used for object detection the later layers should extract information about the position of the object. In the case of object detection, parameter transfer is possible in fewer layers of the neural network.
The usual way of constructing CNNs for detection is to have a feature extractor followed by a detection algorithm borrowing (or not) information from this feature extractor. Hence, before training in an object detection dataset, a good strategy is to initialize the weights to the ones obtained by training the feature extractor to an image recognition dataset (e.g. ImageNet) \cite{56}.

For the purposes of this work, SqueezeDet was selected. We did not choose YOLO\cite{6,7}, which is the state of the art at the moment, because it has many parameters and, as a consequence, it cannot be used in  as many embedded applications as SqueezeDet \cite{9}. The model selection was done after the evaluation of all the major detection networks which were implemented  using Tensorflow \cite{34} and benchmarked using our deep learning workstation\footnote{The workstation architecture includes an intel i7-7700K CPU, a NVIDIA GTX 1080 Ti GPU, 2 $\times$ 8GB RAM @ 2400 MHz, a 100 GB SSD for the code storage and an additional 1TB HDD for the datasets.}.

\subsection{SqueezeDet training}
In the original paper of SqueezeDet, but also in other works \cite{1, 5, 91, 92}, SqueezeDet is trained in the KITTI dataset. It heavilly uses Tensorflow, OpenCV\footnote{\url{https://opencv.org}}, and Numpy \cite{89}. Furthermore, training in KITTI starts with the parameters of the feature extractor trained in ImageNet. As it is presented below, without this prior initialization step the training is impossible.

There is no note of successful SqueezeDet training in other datasets and the number of hyperparameters for training in another dataset increases the search space dimensions to the point that many optimization algorithms do not differ from a random hyperparameter search. There is however a report \cite{3} which implements SqueezeDet with leaky ReLUs, rather than simple ReLUs and it succeeds in training SqueezeDet for all classes of the PASCAL VOC. Also, it splits the data with ratio of 3:1 for training and validation, whereas in our implementation and in SqueezeDet's original implementation the data is split with ratio 1:1 for training and validation.

\subsection{Hyperparameter optimization in CNNs}
\label{subsect:hyperoptimization}
Hyperparameter optimization refers to the hyperparameter space search aiming at the minimization of the model generalization error. The search cannot be done with gradient descent or with a brute force method, because the objective function is a blackbox and any of its samples (e.g. generalization error) is very time consuming to pick.

This broad field of hyperparameter optimization can be approached by many classes of algorithms, like derivative free optimization \cite{84}. To this direction many methods have been developed; some of them are bayesian optimization, TPE, SMBO \cite{86}, genetic algorithms \cite{69}, BOCK \cite{85}, Lipschitz function optimization \cite{87}, etc. The goal of all these methods is to find the global extrema of a function with as few steps as possible avoiding local extrema. One class of methods may not be enough, so libraries containing many types of methods have been implemented, like dlib \cite{83}. In our work, the AdaLipo algorithm \cite{64} was used in combination with a local quadratic model fit around the current best point found, to create a trust region like the one proposed in \cite{65}.

\subsection{Anchor matching}
\label{subsect:AnchorMatching}
Anchor matching is a common technique to match the responsible anchors of the detection layers with the object's shape and position inside the image. It is used in object detection CNNs such as YOLO \cite{6,7}. The problem is that due to data augmentation used in training, responsible anchors have to be computed at each training step. These are usually computed using the CPU and they are an essential part of the neural network's training. This part of the CNN training does not draw too much attention, but it is the bottleneck that does not allow the neural network training modules to reside completely in the GPU.

Essentially, the problem is that of bipartite graph matching. First lets represent the shapes (width, height) and the positions of the objects' centers inside the image as vectors of form $(x, y, w, h)$ and each anchor with indices $[i, j, k]$ respectively with vector $(\hat{x}_i, \hat{y}_j, \hat{w}_k, \hat{h}_k)$. The indices $[i, j, k]$ denote the index in the dimension of the width, the index in the dimension of the height and the index in the dimension of the standard anchor template shapes. For more details, the reader is referred to \cite{1}. We define the distance of the anchors  of the boxes with the one minus the Intersection over Union (IOU) metric.

$$
    1.0 - IOU\left((x, y, w, h), (\hat{x}_i, \hat{y}_j, \hat{w}_k, \hat{h}_k)\right)
$$
Now, let us define a graph $G$ with vertices the anchor and the object vectors, and as edge weights the distances between the anchor and object vectors. The anchors have no edges between them and the boxes have also no edges between them. If $X$ is the vertex set of object vectors and $A$ the vertex set of the anchor vectors, then the bipartite match between $X$ and $A$ is the solution to the anchor matching problem. There are algorithms and fast implementations which solve the problem using the CPU or the GPU \cite{93}.

To the authors' knowledge, there is no object detection framework which solves the problem using the GPU and many networks such as SqueezeDet, YOLO, and others do not provide an exact solution. Their way of solving it is approximating it by doing the steps described below:

\begin{enumerate}
\item They use a dense grid of anchors, such that any wrong matches are not far from the best one.
\item They traverse the objects serially and each time they pick what is best for the current object.
\item They do not allow duplicates - each anchor is assigned to one object.
\end{enumerate}

There are implementations of SqueezeDet's anchor matching for YOLO that overcome this problem and make YOLO trainable in the GPU, but they do not offer the anchor matching as a module that can be used by the community\footnote{\url{https://github.com/nilboy/tensorflow-yolo/tree/python2.7/yolo/net}}. In SqueezeDet training, the anchor matching algorithm is implemented with Numpy and in order to perform backpropagation, information is sent from the GPU to CPU, then the responsible anchors and the error are computed, and the result is sent to the GPU for backpropagation. This process overwhelms the CPU-GPU communication channel. Moreover, very large amounts of data have to be exchanged for this in the SqueezeDet training.

\section{Methodology}
\label{sect:Methods}
\begin{figure}
    \centering
    \includegraphics[width = 0.6\textwidth]{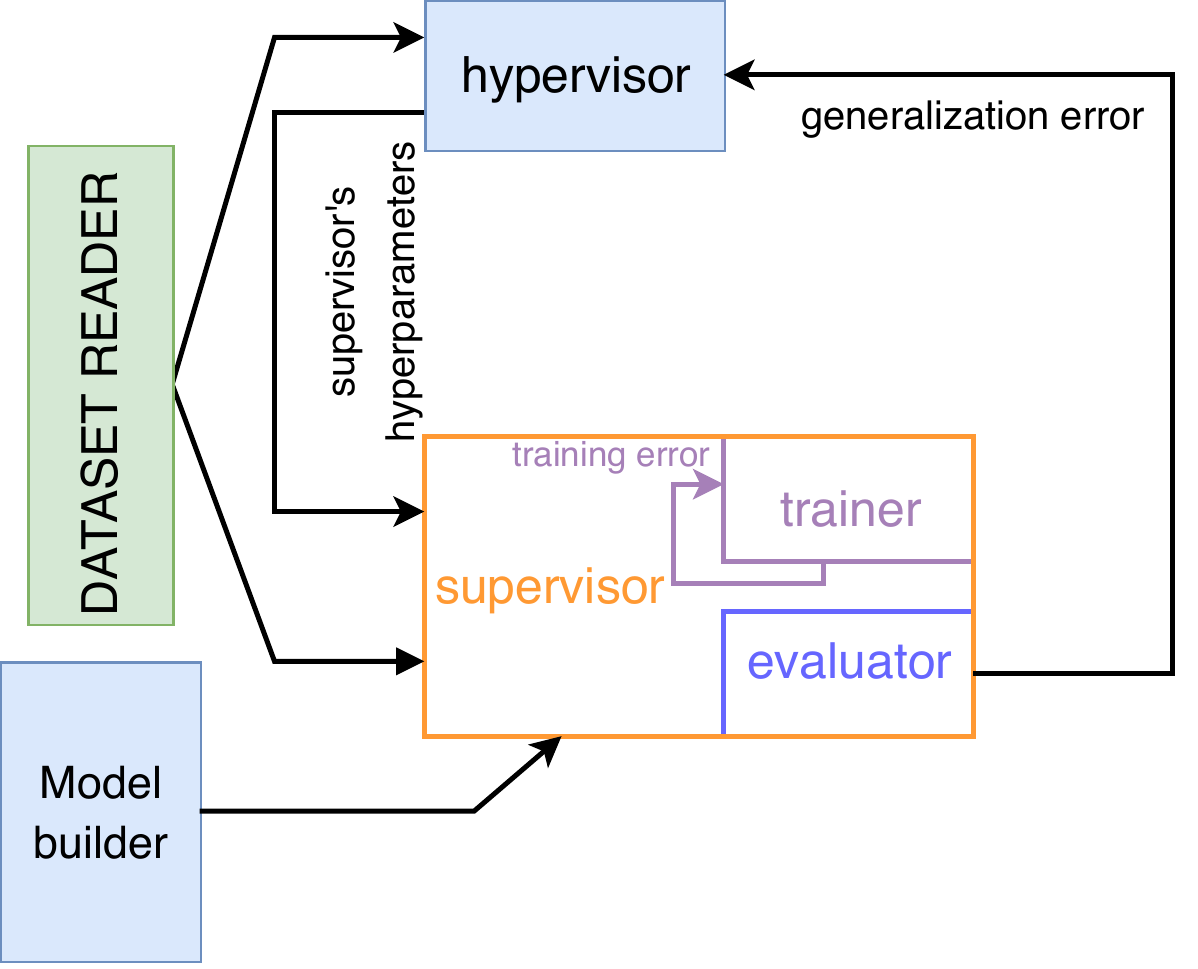}
    \caption[Implementation architecture]{Implementation architecture and information flow. The dataset reader parses the selected dataset using the CPU and sends the parsed data to the hypervisor and the supervisor. The model builder builds the predefined model (e.g. SqueezeDet, SqueezeDet+) to be used by the supervisor. The hypervisor sends the hyperparameters for each training to the supervisor and tries to minimize the generalization error (GE). The supervisor handles the training given the model, the hyperparameters, and the parsed data. It consists of two parts; the evaluator which produces the generalization error and the trainer which trains the model using the training dataset.}
    \label{fig:architecture}
\end{figure}

In this section, we first propose a framework to study the transferability of parameters under hyperparameter optimization. Next, we present a method to analyze the transferability of SqueezeDet, based on the original implementation\footnote{\url{https://github.com/BichenWuUCB/squeezeDet}}. Then, we introduce a way of accelerating the training of object detection CNNs for embedded applications like SqueezeDet. The architecture of this framework is shown in figure \ref{fig:architecture}.

\subsection{Parameter Transfer}
In order to find the transferability of parameters under hyperparameter optimization, the framework is splitted into two parts. The first part performs the steps mentioned in \cite{55} but uses object detection datasets. The second part searches for the optimal hyperparameters that maximize the training process mAP with or without transfer learning. Then, these two are combined to find the optimal hyperparameters and minimize the generalization error (GE) of a retraining which transfers the first $n$ layers of parameters. However, the cost of this optimization is high, so the second part is only called for one of the trainings and then we assume that hyperparameters are optimal for the rest of the trainings. The result is $ min_{hp, n} \left(GE(training(n, hp))\right) $, where $hp$ are the hyperparameters. So, the framework uncovers the optimal procedure for transfer learning automatically without the need of human supervision.

\subsection{SqueezeDet Transferability}

In this study, we use the SqueezeDet network, which uses convDet for detection and as a feature extractor the SqueezeNet CNN. For performing these studies, we use the aforementioned framework. Of course, it is not possible to optimize all the hyperparameters due to the optimization methods' curse of dimensionality, so only a small subset is used and the others are preset manually or produced by a deterministic algorithm depending on the network architecture and the dataset. For selecting the subset of hyperparameters to optimize, before any training, they have to be sorted by their responsibility to the generalization error. Afterwards, the first $N$ hyperparameters, which cannot be computed deterministically, are selected for optimization.

\subsection{Training acceleration}
SqueezeDet training acceleration results from some observations regarding the original implementation, which are listed below:
\begin{enumerate}
\item{Data parsing uses no preprocessed data and requires large amounts of memory because all labels are stored in memory and are processed using CPU multithreading libraries.}
\item{Matrices and tensors are represented in dense form, although the problem's nature allows sparse representations; many trivial computations like $0 \times 0$ could be avoided.}
\item{Image decoding is done using OpenCV and then images are loaded to Tensorflow tensors after data augmentation, which is far from optimal.}
\item{Data augmentation is performed using the CPU, although since it is an embarrassingly parallel procedure it should be better processed by the GPU.}
\item{Anchor matching is performed by the CPU and causes the aforementioned problems.}
\item{Delta computation, which is referred in the SqueezeDet paper, is a completely parallelizable procedure (since the anchors are known), but it is performed using the CPU.}
\item{In the main implementation of SqueezeDet the last filter is implemented using Numpy in CPU which requires large chunks of memory to be transferred from GPU to CPU for filtering. Also, for using the SqueezeDet Tensorflow model with other devices requires that the engineer should write the filtering part again.}
\item{Data visualization uses many libraries requiring more memory.}
\end{enumerate}

To overcome these problems and accelerate the training we minimized the CPU-GPU communication, exchanging only the necessary information and implementing the most procedures on the GPU side using Tensorflow. Particularly, the following steps contributed in achieving the acceleration result:
\begin{enumerate}
\item{Data parsing uses protocol buffers (provided by Tensorflow) which requires constant memory to load all the dataset procedurally and optimally\footnote{\url{https://www.tensorflow.org/performance/datasets_performance}}. This also makes data handling safer. The data are preprocessed and so can be prefetched directly by the GPU during the training.}
\item{Labeling data for object detection is represented as sparse matrix, allowing faster label data transfer separately from the image data.}
\item{Sparse matrix usage on GPU spares trivial computations.}
\item{Data augmentation and image preprocessing is performed by the GPU.}
\item{Anchor matching is performed using the GPU, allowing other parts like pre-processing to be performed also inside the GPU without the need of requesting data from main memory. Analysis of this crucial step is described below.}
\item{Deltas computation is performed by the GPU.}
\item{The final filter is implemented using Tensorflow. This eases data transfer from GPU to CPU. Furthermore, this allows the use of SqueezeDet as a black box without the need of additional code to be written for an embedded device. It is also remarkable that with the use of Tensorflow-Lite\footnote{\url{https://www.tensorflow.org/mobile/tflite/}} SqueezeDet can now be automatically deployed on many devices.}
\item{Visualization uses now only the Tensorflow library and requires less memory.}
\end{enumerate}

\begin{figure}
    \centering
    \includegraphics[height = 0.45\textheight]{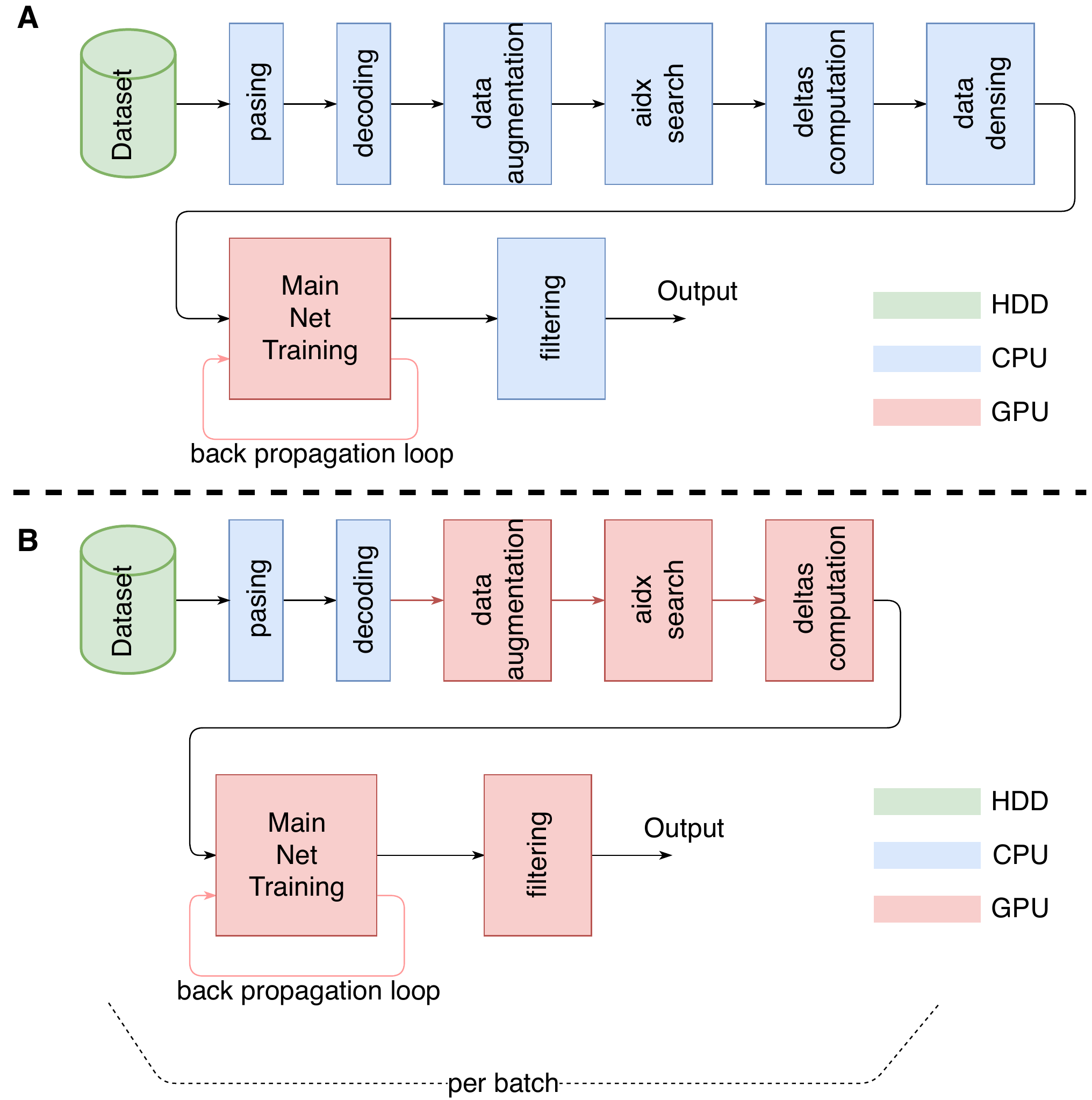}
    \caption[Comparison of the old and new pipelines]{In the old pipeline (A) most stages are computed in the CPU and the GPU is used only for the CNN inference and update stages. In the new pipeline (B), the GPU is used in as many stages as possible. Furthermore, the implementation is based on the Tensorflow framework which allows the model to be imported directly to an application without the need for more code.}
    \label{fig:pipeline}
\end{figure}

These steps can also accelerate the training of other networks. The vital part for these steps is the anchor matching. It allows the label data to be ready inside the GPU and not to be processed by the CPU after data augmentation and finally sent to GPU. The modification introduced in the pipeline is better presented in figure \ref{fig:pipeline}. In this figure, the new pipeline is presented in comparison with the old one.


There are implementations of YOLO which perform the same procedure\footnote{\url{https://github.com/nilboy/tensorflow-yolo/tree/python2.7/yolo/net}}, but they are not directly applicable to other networks. Moreover, most object detection networks use bipartite match algorithms for this step.

\subsection{Anchor matching}
The analysis of parallelizing the anchor matching procedure first requires that the algorithm is written in serial form, as in algorithm \ref{alg:serial}. Then, it is parallelized in algorithm \ref{alg:parallel} which can be written in any GPU API like CUDA. However, because of the use of Tensorflow, this part has to be implemented with linear algebra and matrix equations. This is described in algorithm \ref{alg:Tensor}. 
The steps in algorithm \ref{alg:parallel} do not differ much from algorithm \ref{alg:serial}, but they are presented in a way which shows the parallelizability of each step. Lines $4$ to $5$ in algorithm \ref{alg:parallel} could be computed using a parallel for, rather than compute in parallel each line and continuously synchronize after each step. The same holds for lines $12$ and $13$ and every line inside the $find\_best\_aidx\_per\_image$ function.

\begin{figure}
   \centering
   \includegraphics[width = 0.5\textwidth]{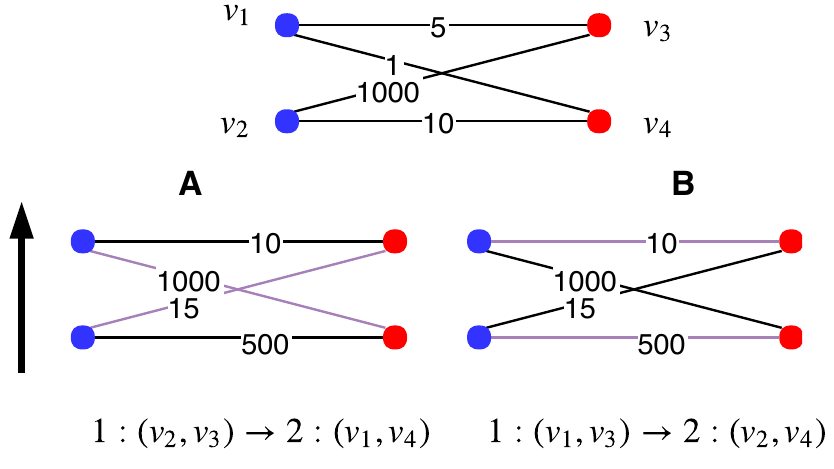}
   \caption[Comparison of two methods of anchor matching]{An example to compare between the result of the algorithm we parallelized (A), in comparison to the result of a greedy bipartite match (B). The black arrow shows the direction of the chosen vertices from the algorithm \ref{alg:serial}. In bottom, we present the order of selection for each case. The algorithm \ref{alg:serial} starts from the vertex $\nu_2$, so it selects first the edge $(\nu_2, \nu_3)$ and then the edge $(\nu_1, \nu_4)$. This produces a total matching weight of $1015$. The greedy bipartite match algorithm first selects the edge with the lower weight: $(\nu_1, \nu_3)$, it removes all edges from $\nu_1, \nu_3$ and then it selects again the edge with the lower weight for the vertices remaining $(\nu_2, \nu_4)$. This produces a total matching weight of $510$. Apparently, the second algorithm has better outcome to this problem. So, to avoid such behaviour it is better to use the second one.}
   \label{fig:RightVSBadAnchorMatching}
\end{figure}

\begin{figure}
   \centering
   \includegraphics[height = 0.45\textheight]{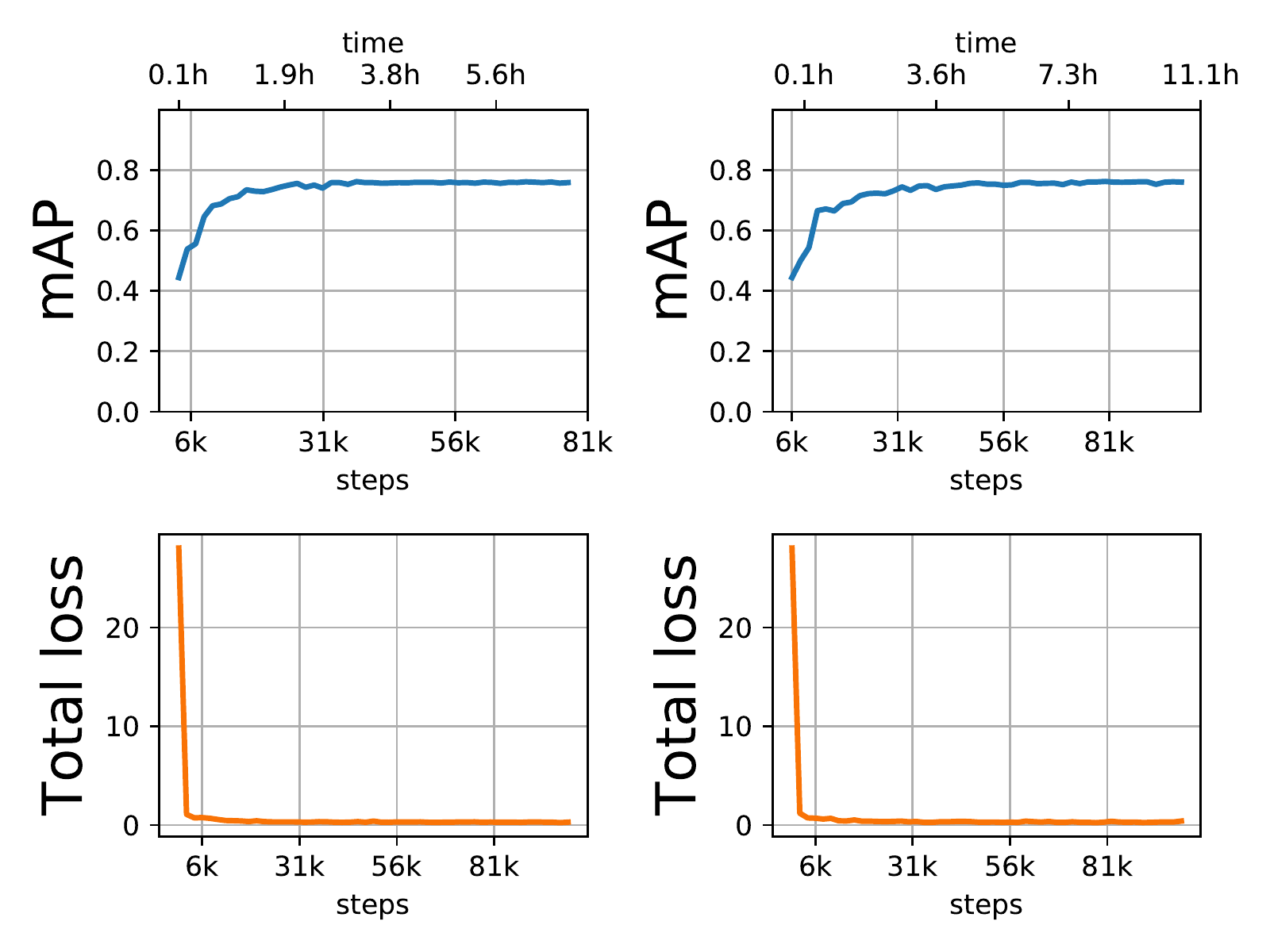}
   \caption[Comparison of two methods of anchor matching in KITTI]{A comparison between the result of the algorithm we parallelized (left) against the result of a greedy bipartite match (right) in the KITTI dataset. The greedy bipartite match is slower than the parallelized one, because we used Tensorflow implementation which is serial and uses only the CPU. So, training suffers from the same problems of memory transfer between CPU and GPU. The final mAP however is the same with both algorithms.}
   \label{fig:kittiCmpAnchorMatching}
\end{figure}

In the second \textbf{for} of algorithm \ref{alg:serial}, is apparent the point where the unconstrained selection of the order of vertices is taking place. Thus, this algorithm is an approximation to the optimal anchor matching. A better approach for solving the same problem, is the greedy bipartite algorithm which is presented in figures \ref{fig:RightVSBadAnchorMatching}, \ref{fig:kittiCmpAnchorMatching}. However, figure \ref{fig:kittiCmpAnchorMatching} shows the training in KITTI, where the two approaches have the same performance.

In algorithm \ref{alg:parallel}, $find\_best\_aidx\_per\_image$ requires the same amount of time as in the serial case, hence the algorithm is not much faster than the serial one. This also holds for algorithm \ref{alg:Tensor}. Nevertheless, the anchor matching alone does not accelerate the training, but allows for other parts to be accelerated.

\begin{algorithm}
   \caption{\textbf{ - Serial form:} In the input, $anchor\_boxes$ is a matrix of anchor boxes of shape $[ANCHORS, 4]$, the boxes form is the one described above. $ground\_truth\_bounding\_boxes$ is a list of length $BATCH\_SIZE$ containing lists with bounding boxes for each image from the dataset. $BATCH\_SIZE$ is the number of images in a batch. Empty lists are defined as $[]$ and empty sets as $\{\}$. Element addition is $\vee$ for lists and $\cup$ for sets.}
   \label{alg:serial}
   \begin{algorithmic}[1]
   \Require anchor\_boxes, ground\_truth\_bounding\_boxes, BATCH\_SIZE
   \State $ anchors\_matched\_to\_images\gets [\{ \}\,for\, 0\ldots BATCH\_SIZE]$
   \For{$idx \in [0,\ldots,BATCH\_SIZE)$}
       \State $ anchors\_used \gets \{\}$
       \For{$bbox \in ground\_truth\_bounding\_boxes$}
           \State $ distances \gets \textrm{compute}\, \left(1-IOU\right)\, \textrm{of bbox with anchor\_boxes}$
           \State $ dist\_ids \gets argsort(distances)$
           \State $ best\_aidx \gets \textrm{find first}\, d\_idx \in dist\_ids%
                   \,|\, distances[d\_idx] < 1, d\_idx \notin anchors\_used$
           \If{$best\_aidx = \textrm{None}$}
                              \State $distances \gets \textrm{euclidean distance of}\, bbox\, \textrm{with}\, anchor\_boxes$

               \State $dist\_ids \gets argsort(distances)$
               \State $best\_aidx \gets \textrm{find first}\, d\_idx \in dist\_ids%
               \,|\, d\_idx \notin anchors\_used$
           \EndIf
           \State $ anchors\_used \gets anchors\_used \cup best\_aidx$
       \EndFor
       \State $anchors\_matched\_to\_images[idx] \gets anchors\_used $
   \EndFor\\
   \Return $anchors\_matched\_to\_images$
   \end{algorithmic}
\end{algorithm}

\begin{algorithm}
    \caption{\textbf{ - Parallel form:} In the input, $rois$ is a sparse matrix which has pointers the elements $rois\_idx$ and values $rois\_values$. It is the Sparse version of $ground\_truth\_bounding\_boxes$. Every line in the algorithm is computed in parallel. The $find\_best\_aidx\_per\_image$ selects aidx (anchors indices) for each image. It finds in every row of matrix $N$, starting from the top, the first element from the "left" different from $-1$. Then it puts $-1$ in every other equal element in $N$ with the element found. $d(.,\,.)$ denotes the euclidean distance.
}
    \label{alg:parallel}
    \begin{algorithmic}[1]
   \Require anchor\_boxes, rois, BATCH\_SIZE, ANCHORS
   \State $num\_rois\gets len(rois_values)$
   \State $R\gets \left[0,\ldots,\,num\_rois\right)$
   \State $B\gets \left[0,\ldots,\,BATCH\_SIZE\right)$
   \State $distances\gets \left(1-IOU(b_1, b_2)\right),\, \forall b_1 \in rois\_values, \forall b_2 \in anchor\_boxes$
   \State $sorted\_dists[i]\gets argsort(distances[i,:]),\, i \in R$
   \State $edist\gets d\left(b_1,b_2\right),\, \forall b_1 \in rois\_values, \forall b_2 \in anchor\_boxes$
   \State $edist\_ids[i]\gets argsort(edist[i,:]),\, i \in R$
   \State $j_i\gets \textrm{find first}\, j\, |\, sorted\_dists[i,j] \leq 0,\,i \in R$
   \State $dist\_ids[i,j_i:]\gets edist\_ids[i,0:ANCHORS-j_i],\,i \in R$\\
   \\%
   // Now get anchor indices corresponding to each image%
   \State $im\_aidx[idx]\gets dist\_ids[i,:]|\,rois\_idx[i,0]=idx,\,idx\in B$
   \State $aidx\_values\gets\bigvee_{i=0}^{BATCH\_SIZE-1} find\_best\_aidx\_per\_image(im\_aidx[i])$
   \State $anchors\_to\_images\gets SparseArray(rois\_idx, aidx\_values)$\\
   \Return $anchors\_to\_images$
   \\

    \Function{$find\_best\_aidx\_per\_image$}{$aidx\_slice$}
        \State $slice\_shape\gets shape(aidx\_slice)$
        \State $els\_used\gets [aidx\_slice[0][0], 0,\ldots,0]$
        \State $N\gets aidx\_slice$
        \State $i\gets 1$
        \While{$i < len(aidx\_slice)$}
            \For{$j \in [0,\ldots,slice\_shape[1])$} \textbf{in parallel}
                \If{$N[i][j] = els\_used[i-1]$}
                    \State $N[i][j]\gets -1$
                \EndIf
            \EndFor
            \State $s\gets \bigvee_{j=0}^{slice\_shape[1]-1} N[i][j]\,|\,N[i][j] \neq -1$
            \State $els\_used\gets s[0]$
            \State $i\gets i+1$
        \EndWhile\\
        \Return $els\_used$    
    \EndFunction
    \end{algorithmic}
    
\end{algorithm}

\begin{algorithm}
    \caption{\textbf{ - Tensorflow form:} The only adjustment to algorithm \ref{alg:parallel} is inside $find\_best\_aidx\_per\_image$. The $\mathbf{while}$ computes everything concurrently. The reason is that the Tensorflow framework requires both iteration index $i$ and the iteration body to be executed simultaneously.}
    \label{alg:Tensor}
    \begin{algorithmic}[1]
   \Require anchor\_boxes, rois, BATCH\_SIZE, ANCHORS

    Main part is the same as the simple parallel algorithm.
    \Function{$find\_best\_aidx\_per\_image$}{$aidx\_slice$}
        \State $slice\_shape\gets shape(aidx\_slice)$
        \State $els\_used\gets [aidx\_slice[0][0], 0,\ldots,0]$
        \State $neg\_ones\gets -\mathbf{1}_{i,j}, (i,j) \in \left[(0,0),\ldots,(shape(aidx\_slice)-(1,1))\right]$
        \State $N\gets aidx\_slice$
        \State $i\gets 1$
        \State $J\gets [0,\ldots,slice\_shape[1])$
        \While{$i < len(aidx\_slice)$}
            \State $N[i][j] \gets -1 \; \textstyle{if}\; N[i][j] = els\_used[i-1] \; \textstyle{else}\; N[i][j], j \in J$
            \State $els\_used\gets [els\_used[0],\ldots,els\_used[i-1],N[i,min(j|(N[i][j]\neq-1\, and \, N[i][j]\neq els\_used[i-1]))],0,\ldots,0] $
            \State $i\gets i+1$
        \EndWhile\\
        \Return $els\_used$    
    \EndFunction
    
    \end{algorithmic}
    
\end{algorithm}

The anchor matching used in SqueezeDet follows the strategy described in section \ref{subsect:AnchorMatching} and its simple form is presented in algorithm \ref{alg:serial}. But, this algorithm's performance depends on the grid's high density. If the grid is thick, then any box that is close to the object is close to the responsible anchor. That does not mean that the training's results will always be the same, because we introduce noise that should not be there. In figure \ref{fig:RightVSBadAnchorMatching}, a false outcome of this method is presented in contrast to the outcome of a greedy bipartite match method. The greedy bipartite match instead works with the following three steps:

\begin{enumerate}
   \item Sort all edges using as key their weights to an array.
   \item Choose the first $n$ smallest edges from the sorted array.
   \item When choosing an edge, take care that its nodes have not been selected  for any previously chosen edge.
\end{enumerate}

Although these three steps should provide better performance as seen in figure \ref{fig:RightVSBadAnchorMatching}, in figure \ref{fig:kittiCmpAnchorMatching} we show that the result is the same. Also, the algorithm implementing these steps is harder to parallelize than our parallelization approach above. It requires sorting hundreds of thousands of edges and keeping track of their nodes in parallel. But, it can work better in more sparse grids. So, we leave this as future work.

\subsection{Hyperoptimization}
In order to choose the right hyperparameters, many steps should be taken. But, for a few hyperparameters, the researcher or the application engineer has no clue which value to use; she or he knows only the range of the values a hyperparameter could take. To solve this issue we used a hyperoptimization method as we mentioned in section \ref{subsect:hyperoptimization}. The use of the combination of AdaLipo and a trust region method is empirical. The Adalipo falls into the Lipschitz function optimization class. This algorithm models the training process as a black box. There is no official report till now for using AdaLipo in neural network hyperparameter optimization, but neural networks and CNNs are Lipschitz functions \cite{88} and so this algorithm is applicable to them. But, even if some hyperparameters cause them not to have this property, then it will be like using only the trust region algorithm combined with a random search.

\begin{figure}
    \centering
    \includegraphics[height = 0.45\textheight]{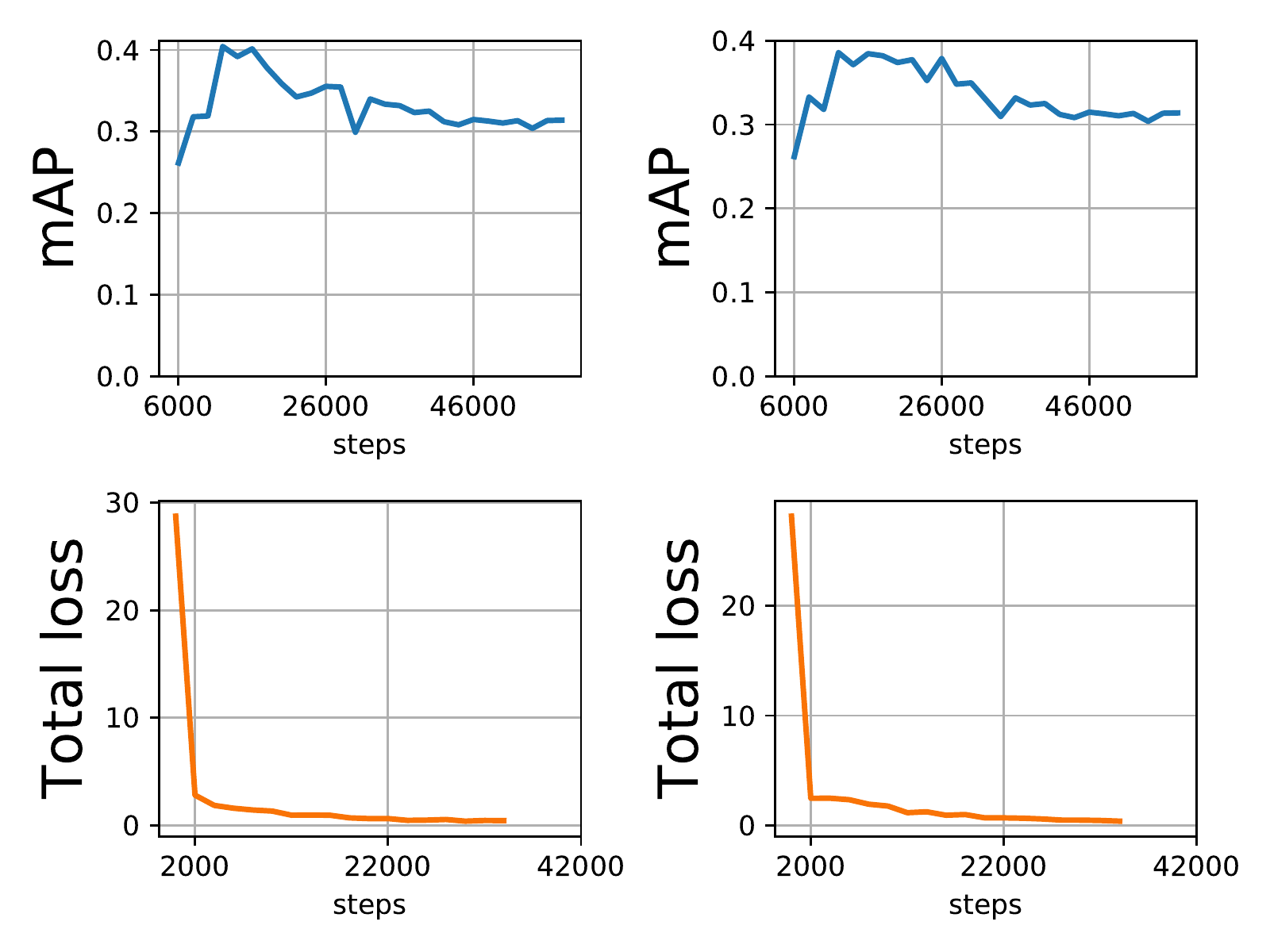}
    \caption[Starting SqueezeDet training in PASCAL VOC with poor input]{Starting SqueezeDet training in PASCAL VOC with poor input. On the right side of the figure a retraining using ImageNet weights is presented, where SqueezeDet does not generalize well due to poor input; in this case it does not achieve mAP over $0.4$. However, the SqueezeDet generalizes better in another retraining using the same hyperamarameters, as shown at the left of the figure. Also, below are presented the total losses of the unsuccessful and successful trainings, right and left respectively.}
    \label{fig:poorVSgoodPascal}
\end{figure}

The need for a trust region method comes from the fact that there is noise in the input. If we retrain for PASCAL VOC for 40000 steps, it is not guaranteed that we will have exactly the same results (mAP) in the evaluation dataset in another training with the same hyperparameters. This is due to two facts. First, the input is always in random order. This can be seen in figure \ref{fig:poorVSgoodPascal}. Second, we sample the evaluation at a number of steps and not  in every step, because it is time consuming. The hyperoptimization method models this result inconsistency as white noise.

\section{Experiments}
\label{sect:Experiments}
\subsection{Comparing CNNs}
Before advancing to the experiments of Transfer Learning we would like to have an idea about the speed of each network. First in table \ref{table:cnnComp}, a comparison between feature extractors is presented. For the comparison, we added functionality to the official Tensorflow repository for performance measurements\footnote{\url{https://github.com/supernlogn/benchmarks}}. Tables \ref{table:cnnComp} and \ref{table:detCnnComp} justify our selection of the SqueezeNet CNN as the most suitable for use in embedded systems.

\begin{table}
    \centering
    
    \begin{tabular}{c|c|c|c}
     feature extractor & Top-1 accuracy & \#Parameters & frames/ s /W \\
    \hline
    
    SqueezeNet \cite{2} & 57.5 & 1.4M& $ 9.429 $ \\ 
    XNOR-Net \cite{18} & 44.2 & 61M & $ 219.78 $ \\ %
    VGG-16 & 70.5 & 14.7M& $2.225$ \\ 
    MobileNet-224 & 83.3 & 3.2M & $45.278$ \\ %
    ResNet-101 V2 \cite{21} & 76.4 & 42.6M & $2.316$ \\ 
    Inception V3 & 78.0 & 21.8M & $ 2.4276$ \\  
    Inception V4 & 80.4 & 54.3M & $ 1.3816$ \\ 
    \hline
    \end{tabular}
    \caption[Feature extraction networks comparison]{Comparing networks for feature extraction in forward pass per frame. All parameters are represented in float32 except from the XNOR-Net parameters which are binary. XNOR-Net executes in CPU and a different platform (\textit{ATOM Z530} instead of our GTX 1080 Ti based deep learning workstation). Accuracy is measured in ImageNet. Network characteristics are taken from \cite{17, 28, 29, 30, 31, 32}. Image shape is the same as in ImageNet.}
    \label{table:cnnComp}
\end{table}

\begin{table}
    \begin{tabular}{c|c|c}
     Detection CNN & mAP & Exec Time / frame / W \\
    \hline
    SqueezeDet & 80.4 (KITTI) &$ 0.136\,ms / W$ \cite{1} \\  %
    YOLO9000 $480\times480$ + Darknet & 77.8 (PASCAL VOC 7+12) & $ 0.068\,ms / W$ \cite{7} \\ %
    SSD300 + VGG16 & 74.3 (PASCAL VOC 7+12) & $ 0.208\,ms / W$ \cite{26} \\ %
    {Mask R-CNN + ResNeXt-101} \textcolor{red}{*} & 37.1 (MS COCO 2015) & $ \approx 1.677\,ms / W$ \cite{13} \\ %
    {Faster R-CNN + ResNet} & 76.4 (PASCAL VOC 7+12) & $1.397\,ms / W$ \cite{12}\\%
    \hline
    \end{tabular}
    \caption[Object detection networks comparison]{Object detection networks comparison. Accuracy is not measured in the same dataset, since there is no common dataset for all, but all nets can adapt in different object detection sets. All results are from computations with NVIDIA Titan X GPU except from (\textcolor{red}{*}) which used NVIDIA Tesla M40 GPU.}
    \label{table:detCnnComp}
\end{table}

\subsection{Transfer Learning between datasets}
In the experiments we followed the notation used in \cite{55}. KITTI is denoted as dataset $A$ and a subset of PASCAL VOC as dataset $B$. Dataset $A$ contains the classes "car, pedestrian, cyclist" as in the SqueezeDet paper. For dataset $B$, we tried different subsets of PASCAL VOC classes; here is presented the subset "bicycle, bus, dog" which is relevant to dataset $A$.

\begin{figure}
    \centering
    \includegraphics[height = 0.45\textheight]{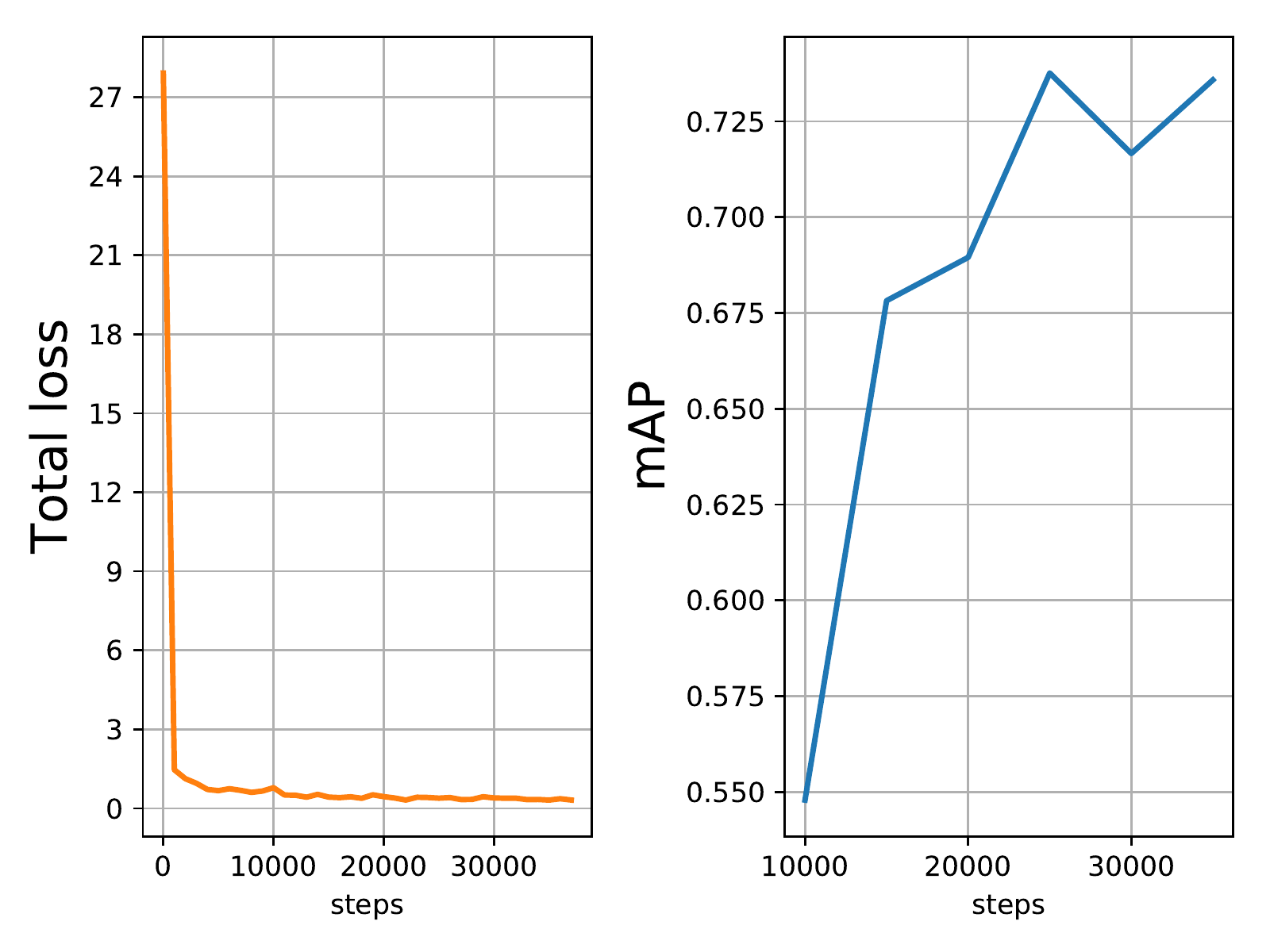}
    \caption[SqueezeDet training in dataset A, with use of parameters from training in ImageNet]{SqueezeDet training in dataset $A$, with using feature extractor parameters from training in ImageNet. Parameter transfer from ImageNet. Training is presented only for the first 45000 steps.}
    \label{fig:Atrain}
\end{figure}

\begin{figure}
    \centering
    \includegraphics[height = 0.35\textheight]{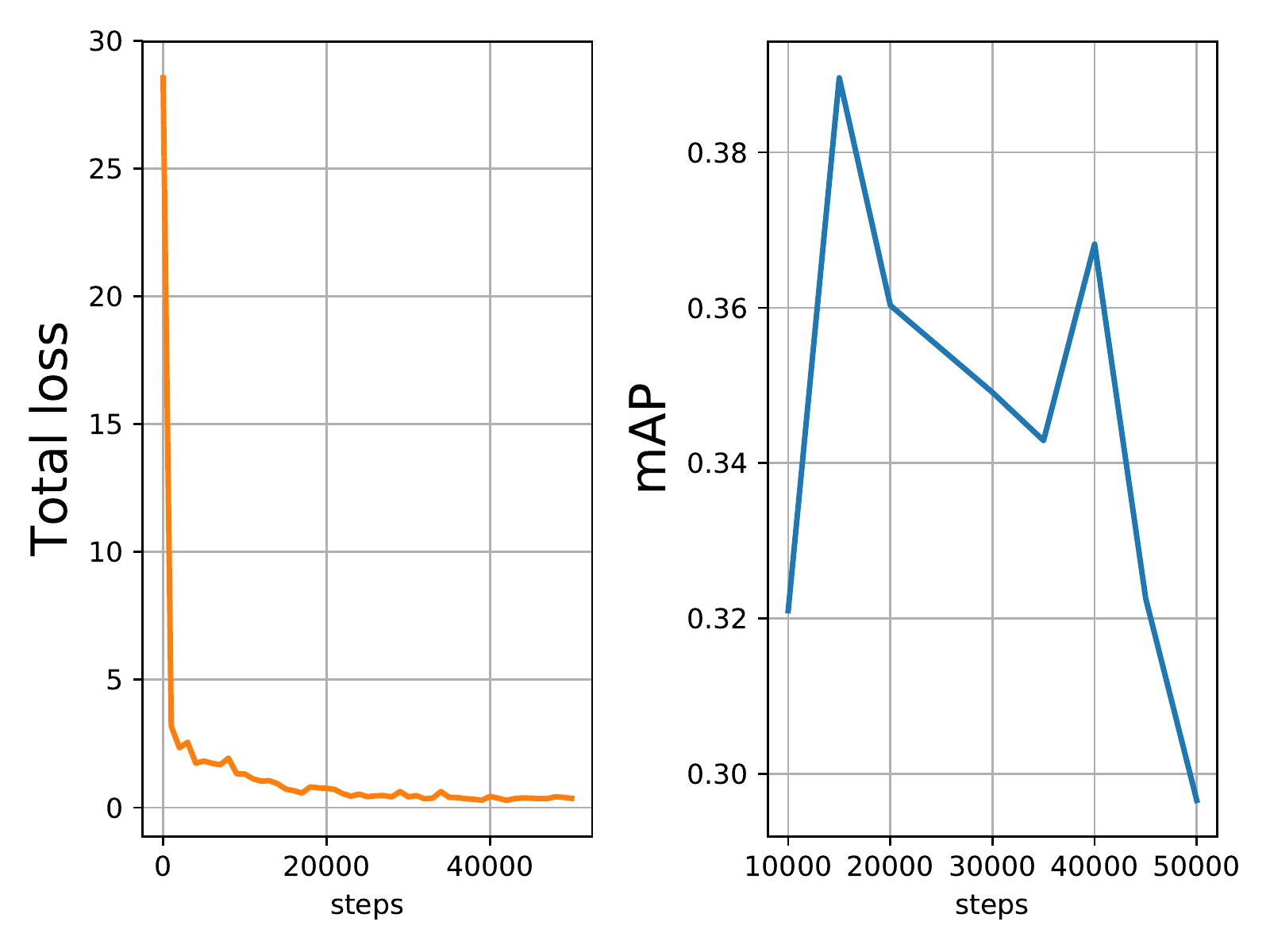}
    \caption[SqueezeDet training in dataset B, with use of parameters from training in ImageNet]{SqueezeDet training in dataset $B$, with using parameters from training in ImageNet. This is the usual way and recommended for training object detection networks to small datasets.}
    \label{fig:Btrain1}
\end{figure}

\begin{figure}
    \centering
    \includegraphics[height = 0.35\textheight]{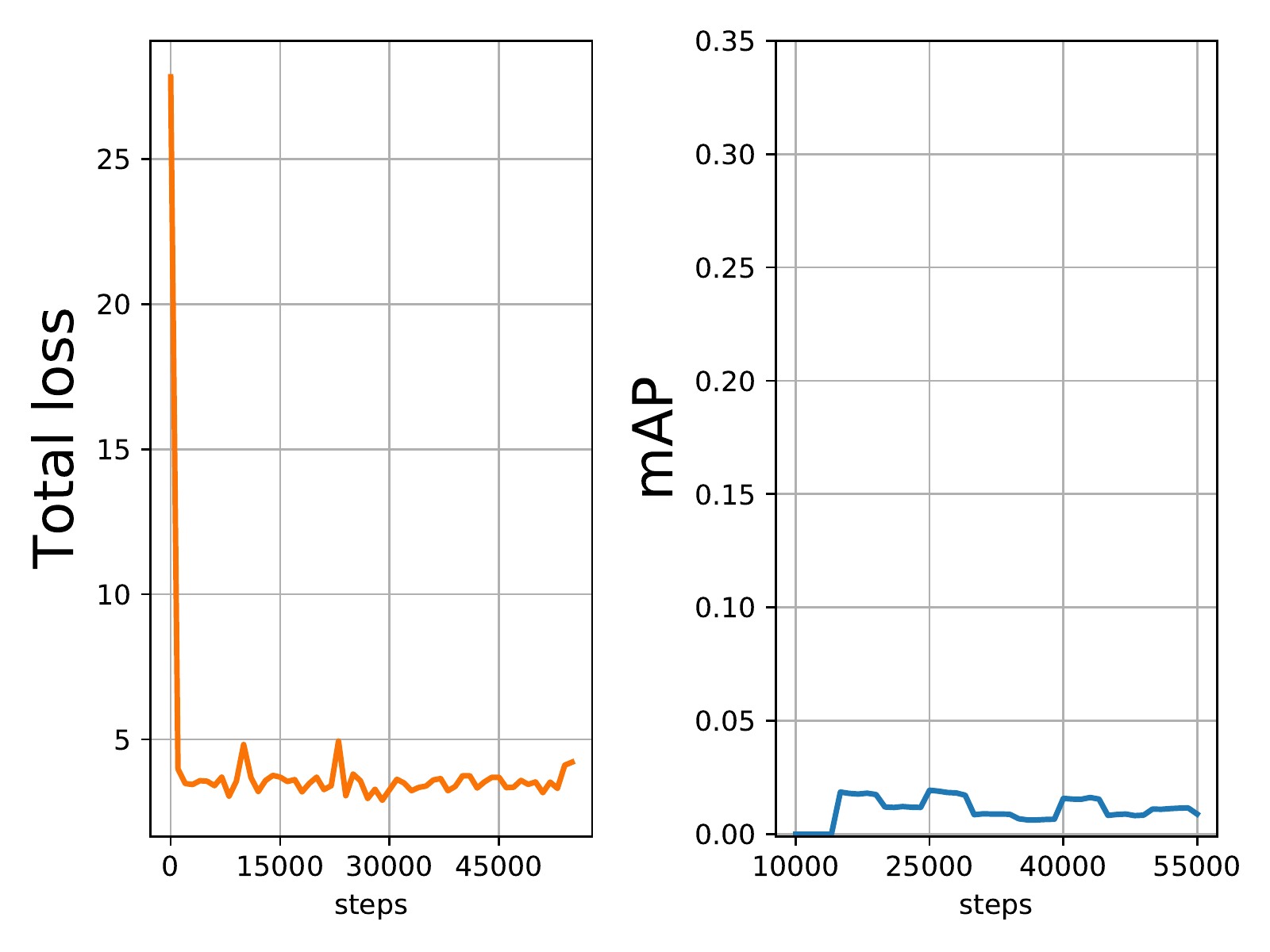}
    \caption[SqueezeDet training in dataset B, without using parameters from training in ImageNet]{SqueezeDet training in dataset $B$, without using parameters from training in ImageNet. Training for $55000$ iterations does not produce any improvement in mAP. For a sufficient mAP more data is required.}
    \label{fig:Btrain2}
\end{figure}

\begin{figure}
    \centering
    \includegraphics[height = 0.35\textheight]{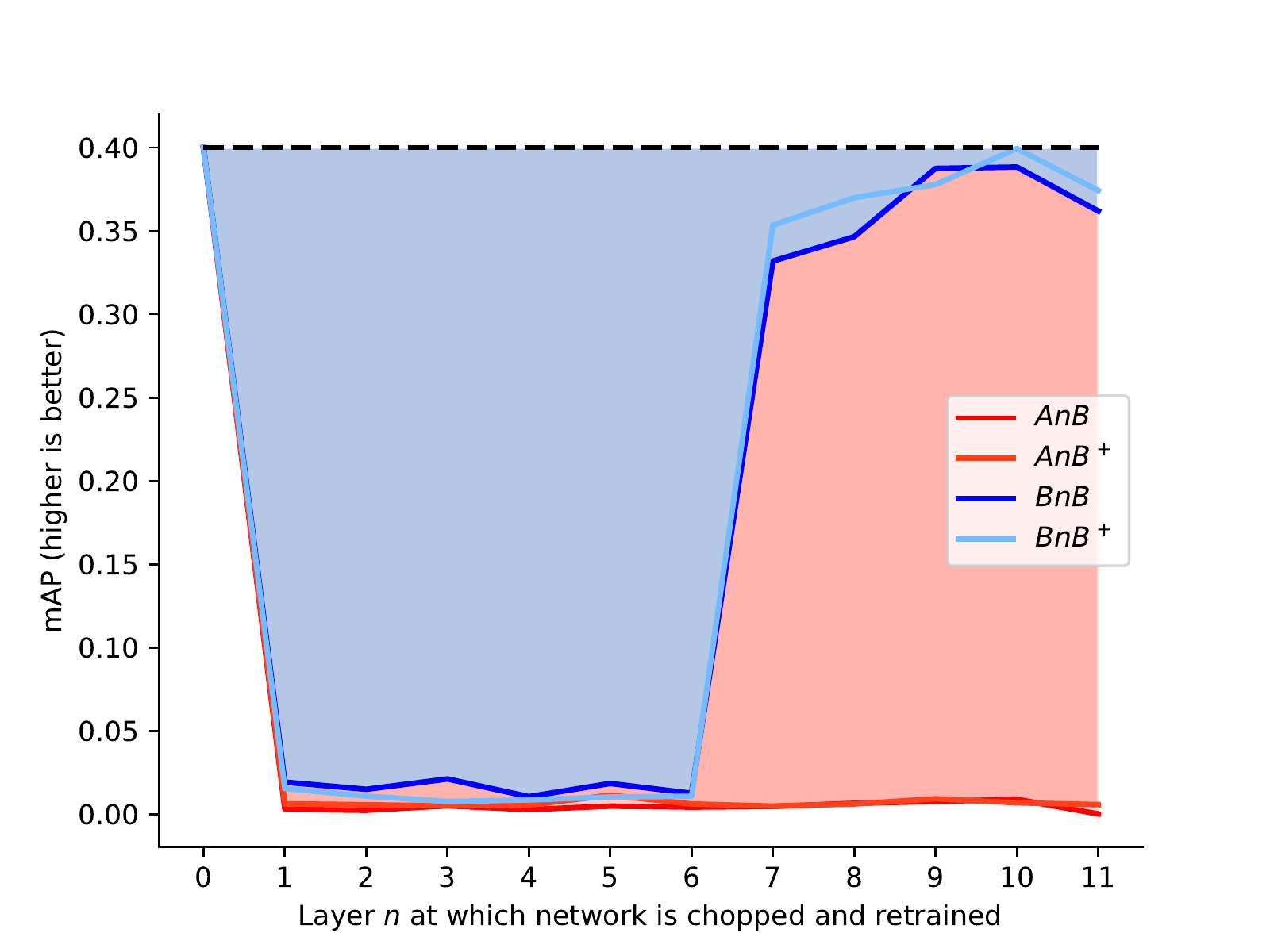}
    \caption[SqueezeDet transferability experiment from dataset A to dataset B]{SqueezeDet transferability experiment from dataset $A$ to dataset $B$. Blue line corresponds to the mAP of $BnB$, the soft blue to $BnB^+$ , the red to $AnB$ and the soft red to $AnB^+$.}
    \label{fig:grantExpRes}
\end{figure}

\begin{figure}
    \centering
    \includegraphics[height = 0.35\textheight]{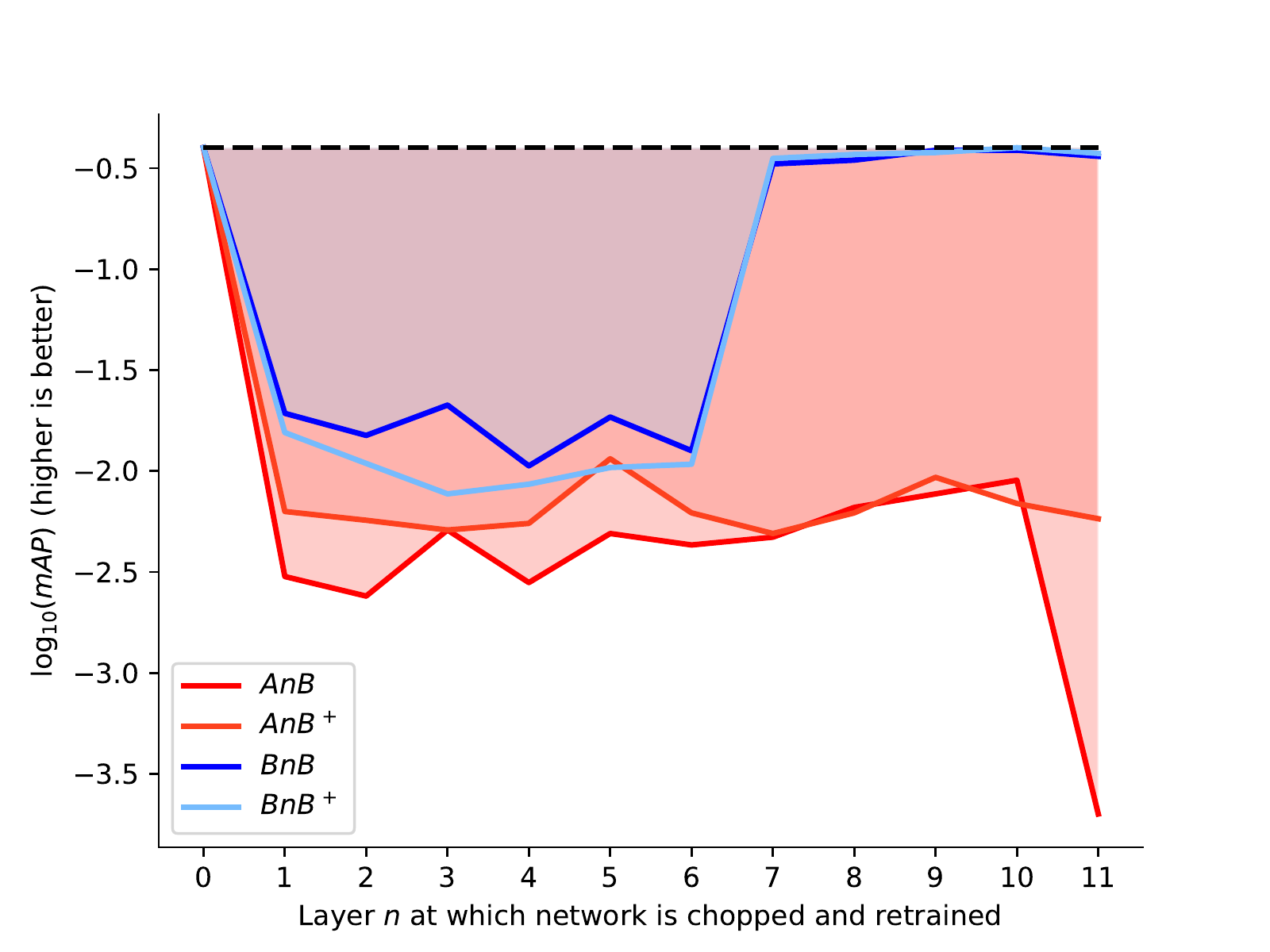}
    \caption[SqueezeDet transferability experiment from dataset A to dataset B in logarithmic scale]{SqueezeDet transferability experiment from dataset $A$ to dataset $B$ in logarithmic scale. Blue line corresponds to the mAP of $BnB$, the soft blue to $BnB^+$ , the red to $AnB$ and the soft red to $AnB^+$. The difference between $BnB$ and $BnB^+$ is more distinct in this diagram.}
    \label{fig:grantExpResLog}
\end{figure}

An elementary problem arises when conducting these experiments. In object detection, it is a common strategy to use transfer learning. So, most algorithms, especially those used for embedded devices, have been developed to be trained using transfer learning. For the training in dataset $A$, as is presented in figure \ref{fig:Atrain}, that does not cause a problem. But for dataset $B$, it is apparent from figures \ref{fig:Btrain1} and \ref{fig:Btrain2} that without transfer learning the training fails. In the transferability experiment in figures \ref{fig:grantExpRes}, \ref{fig:grantExpResLog} this causes the parameter transfer from dataset $A$ to $B$ to fail and so the transferability cannot be studied. The same occurs even if the parameters are fine tuned. To verify that this was not an error of our implementation, we have also tested trainings for 100000 iterations ($=1800\, epochs$) and for different hyperparameter values. Therefore, we assume that the negative transferability is caused only by the insufficient data.

However, the studies of fragility to co-adaptation could be performed. Furthermore, the retraining of $BnB$ and $BnB^{+}$ were tested for $n = 4,5$ for $65000$ steps and the results were the same. Hence, we conclude that retraining with more steps does not produce better mAP. $BnB$ retraining seems to perform worse in the verification dataset than the initial training $B$. This occurs because of overfitting in the training dataset $B$. If $BnB^{+}$ retraining had greater mAP than $BnB$, then the behaviour of retraining $AnB^{+}$ would be better than $AnB$, but, $BnB^{+}$ has lower mAP.

\subsection{Acceleration measurements}
\begin{figure}
    \centering
    \includegraphics[height = 0.4\textheight]{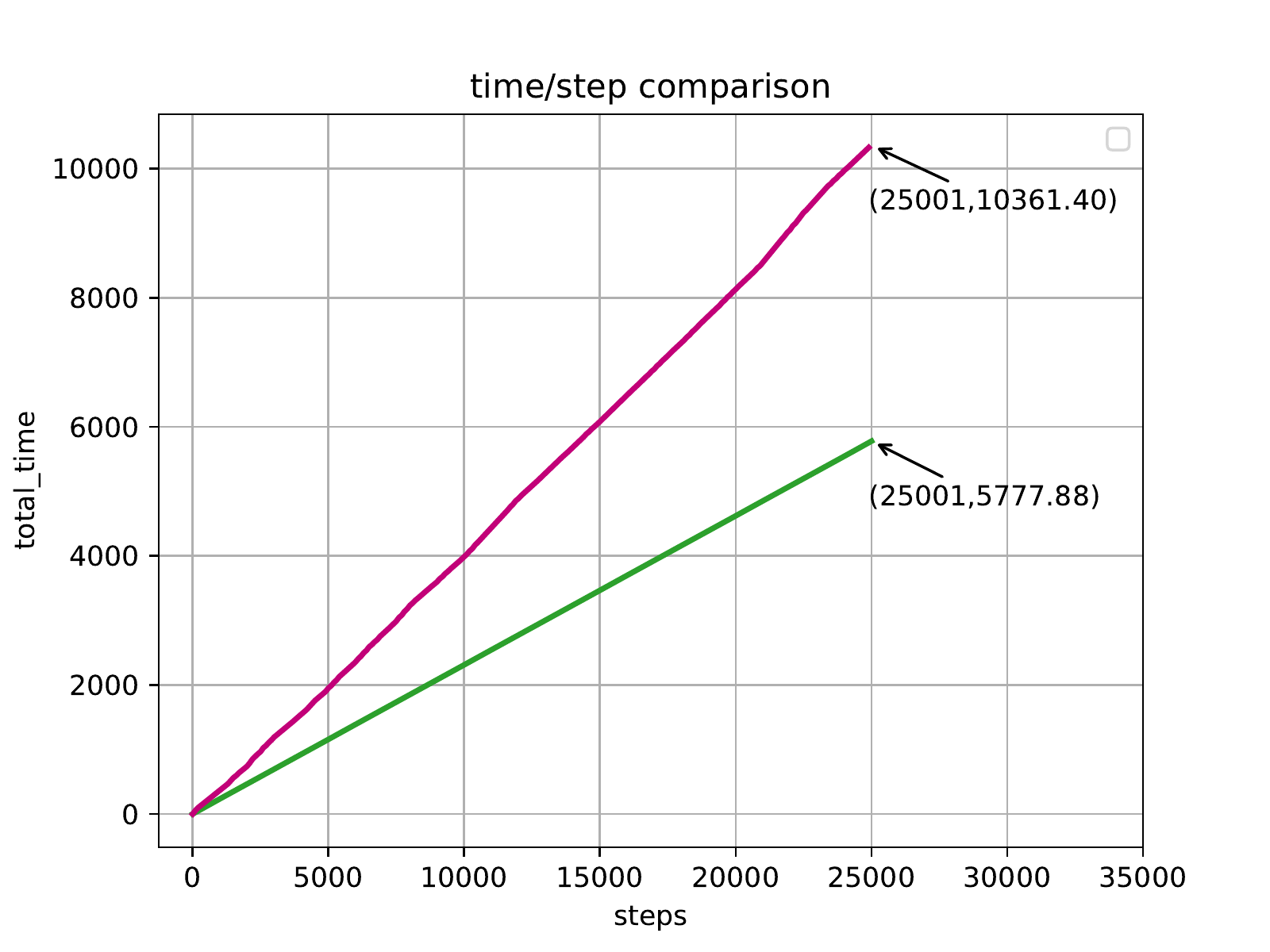}
    \caption[Training time comparison of new and old implementations]{Training time comparison of new and old implementations. Time is measured in seconds. Acceleration equals $1.799 \approx 1.8$. \textcolor{purple}{Purple} corresponds to the old implementation and \textcolor{SeaGreen}{Green} to the new.}
    \label{fig:kittiCompare}
\end{figure}

\begin{figure}
    \centering
    \includegraphics[height = 0.4\textheight]{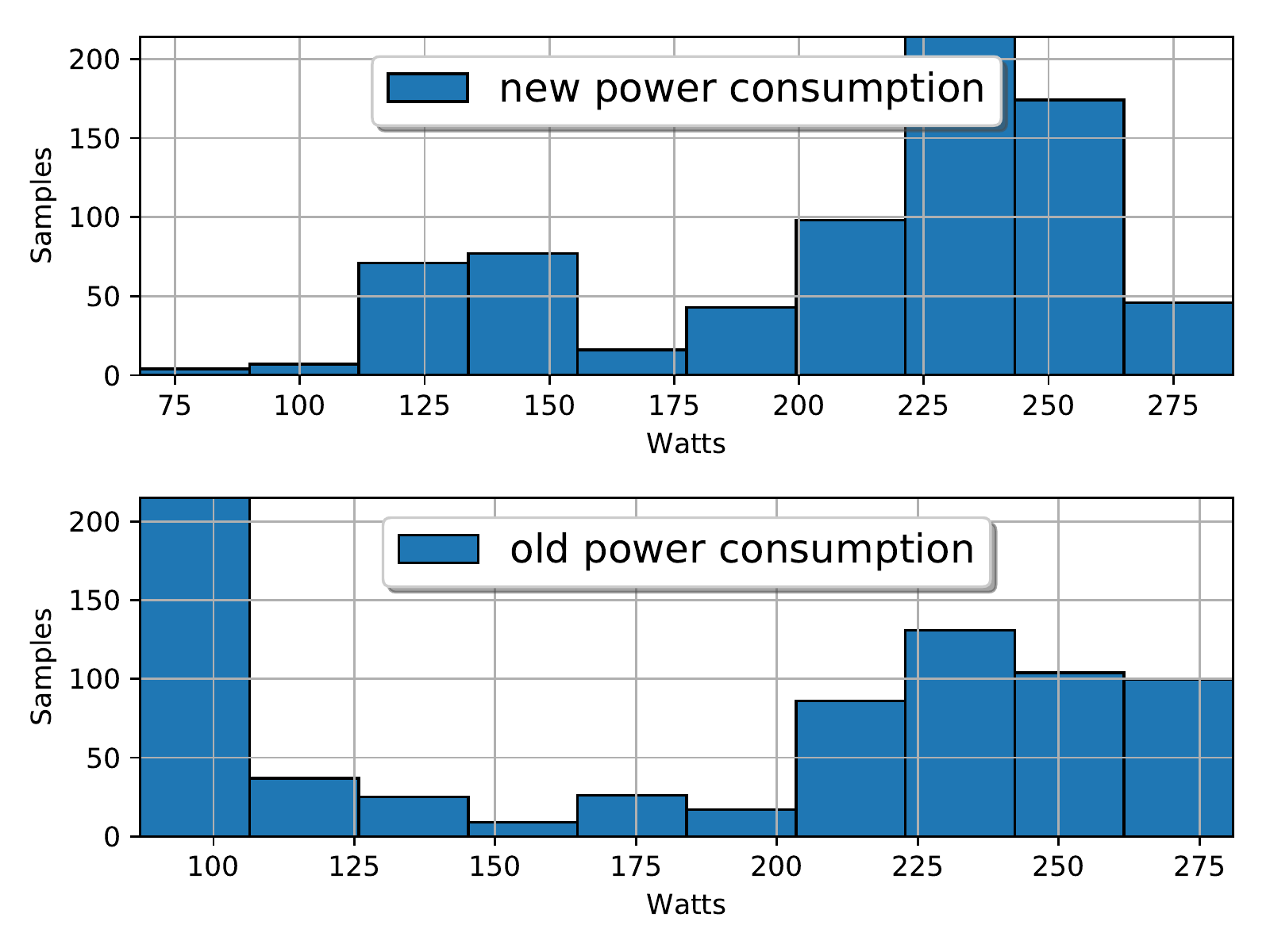}
    \caption[Power consumption comparison of new and old implementations]{GPU power consumption comparison of new and old implementations. It is evident that the new implementation uses more power than the old one. So it performs more GPU computations, which is desirable.}
    \label{fig:powerCompare}
\end{figure}

The acceleration approach has increased the speed of training by a factor of $1.8 \times$. The result of acceleration in KITTI for the first $20000$ steps can be seen in figure \ref{fig:kittiCompare}. The new implementation was also verified for correctness in the KITTI dataset. It is remarkable that in PASCAL VOC the speed of processing reaches 11 $batches/sec$ which is equal to $220$ images of shape $(334, 500)$ per second. This means that an epoch which is 386 iterations in PASCAL VOC requires 89 $sec$. This is a result of minimum data transfer and higher GPU usage. The later can be seen in figure \ref{fig:powerCompare}, for which we altered the official Tensorflow benchmarks repository to measure power consumption concurrently with execution and time measurements.

During the transferability experiment the hyperparameters were constant, but to pick them first we used an optimization approach for the retraining of the case defined by $AnB+, n = 10$. Since we cannot optimize every hyperparameter, we picked only the ones that are more relevant and can not be optimized deterministically. In table \ref{table:hypSearch} we show the chosen hyperparameters to be optimized and their final value. The hyperoptimization iterations were 70 using the aforementioned method. The best mAP was 1.3\% with random selection and with optimization $mAP=2.8\%$. Hence, even under hyperparameter optimization the retraining could not produce better results.

\begin{table}[]
    \centering
    \caption{Hyperparameter optimization search space and optimal values after $70$ steps}
    \label{table:hypSearch}
    \begin{tabular}{|l|l|l|l|}
    \hline
    \textbf{hyperamarameter} & min & max & optimal value\\ \hline
    \textit{ANCHOR\_PER\_GRID} &  1 & 16 & 15\\ \hline
    \textit{NMS\_THRESH} &  0.0 & 1.0 & 0.487\\ \hline
    \textit{LEARNING\_RATE} & 0.01 &  0.10 & 0.01\\ \hline
    \textit{WEIGHT\_DECAY} & 0.00001 & 0.00100 & 0.000521\\ \hline
    \end{tabular}
\end{table}

\section{Conclusion \& Future Work}
\label{sect:Conclusions}
In this paper, we introduced a framework for transfer learning in CNN architectures for object detection. Specifically, this framework allowed us to do transferability experiments and study the fragility to co-adaptation phenomenon. To accelerate the training process, we re-implemented the SqueezeDet's training pipeline speeding it up by a factor of $1.8\times$. For achieving automatic hyperparameter optimization, we used a new empirical method combining AdaLipo with a trust region algorithm.

The transferability experiment from KITTI to PASCAL VOC was unsuccessful, but it successfully presented how fragile to co-adaptation a CNN for embedded applications is. The results showed that at least the first $7$ layers of SqueezeDet have to be fine tuned during a retraining. This may be caused by the use of small dataset; studying this has to go beyond the single case of PASCAL VOC dataset and the SqueezeDet model.

Moreover, the anchor matching algorithm which caused large memory transfers, slowing down the initial SqueezeDet training process, was parallelized in a form capable for GPU to execute. We studied this algorithm for its validity and presented that is unable to solve some problems introducing noise to the training. So, we tried another common algorithm used for anchor matching, greedy bipartite match, which took more time to produce the same results, because the anchor grid was thick. Further experiments comparing these two anchor matching approaches in both time and accuracy could be performed in the case of a more sparse anchor grid.

Furthermore, hyperparameter optimization did not produce greater results than the random ones, so wrong hyperparameters are not a subject into question. Although transfer learning from ImageNet to KITTI and transfer learning from ImageNet to PASCAL VOC is successful, transfer learning from ImageNet to KITTI and then to PASCAL VOC fails. This is an example of catastrophic forgetting \cite{62}.

\bibliographystyle{elsarticle-num}
\bibliography{references}
\vspace{2em}




\end{document}